# Title
Harnessing Smartphone Sensors for Enhanced Road Safety: A Comprehensive Dataset and Review

# Authors
Amith Khandakar[1], David G. Michelson[2], Mansura Naznine[3], Abdus Salam[1,4], Md. Nahiduzzaman[4], Khaled M. Khan[5], Ponnuthurai Nagaratnam Suganthan[6], Mohamed Arselene Ayari[1], Hamid Menouar[7], Julfikar Haider[8]

**Affiliations**
1. Department of Electrical Engineering, Qatar University, Doha 2713, Qatar.
2. Department of Electrical and Computer Engineering, The University of British Columbia, Vancouver, BC V6T 1Z4, Canada
3. Department of Computer Science & Engineering, Rajshahi University of Engineering & Technology, Rajshahi 6204, Bangladesh.
4. Department of Electrical & Computer Engineering, Rajshahi University of Engineering & Technology, Rajshahi 6204, Bangladesh.
5. Department of Computer Science and Engineering, Qatar University, Doha 2713, Qatar
6. KINDI Center for Computing Research, College of Engineering, Qatar University, Doha, Qatar
7. Qatar Mobility Innovations Center (QMIC), Qatar University, Doha, Qatar
8. Department of Engineering, Manchester Metropolitan University, Chester Street, Manchester M1 5GD, UK (United Kingdom)

**Corresponding author:**
Amith Khandakar
Department of Electrical Engineering
Qatar University
Doha 2713, Qatar
E-Mail: amitk@qu.edu.qa

## Abstract

Severe collisions can result from aggressive driving and poor road conditions, emphasizing the need for effective monitoring to ensure safety. Smartphones, with their array of built-in sensors, offer a practical and affordable solution for road-sensing. However, the lack of reliable, standardized datasets has hindered progress in assessing road conditions and driving patterns. This study addresses this gap by introducing a comprehensive dataset derived from smartphone sensors, which surpasses existing datasets by incorporating a diverse range of sensors including accelerometer, gyroscope, magnetometer, GPS, gravity, orientation, and uncalibrated sensors. These sensors capture extensive parameters such as acceleration force, gravitation, rotation rate, magnetic field strength, and vehicle speed, providing a detailed understanding of road conditions and driving behaviors. The dataset is designed to enhance road safety, infrastructure maintenance, traffic management, and urban planning. By making this dataset available to the community, the study aims to foster collaboration, inspire further research, and facilitate the development of innovative solutions in intelligent transportation systems.

## Background & Summary

According to statistics, worldwide car usage is expected to almost double by 2040 [1]. More vehicles require exploring new technologies and methods to guarantee road safety, including driver assistance systems, driver monitoring devices, and driver training programs. The safety of drivers, passengers, and pedestrians can be threatened by aggressive and unpredictable driving, as well as by traffic conditions (slippery roads, obstacles on the road, mudslides, animals on the road, etc.). [2] Researchers observed more honking, rude gestures, and unsafe lane changes by drivers on roads with more potholes and other anomalies because these roads require more attention and vehicle control, which can make drivers angry. Other factors that contribute to this risk include fatigue, alcohol consumption, and the use of a cell phone while driving [3–6]. Driving behaviors vary from person to person; a person's character and habits influence how they will respond in a potentially risky circumstance while operating a vehicle [7,8]. An aggressive driver adopts a driving style that combines high speeds with frequent and abrupt changes in instantaneous speed, resulting in abrupt accelerations and decelerations. In recent years, there has been a significant focus on driving behavior analytics as a crucial tool to improve road safety [9–11].

High-quality road infrastructure ensures efficient movement of people, goods, and facilities, as well as access to a variety of economic and social activities. The consequences of climate change, city expansion, population growth, and improved road design are some of the reasons that can cause transportation infrastructure to rapidly deteriorate. Road anomalies such as potholes, cracks, and bumps are common due to this degradation and can be observed on roads throughout the world [12]. Untreated pavement cracks can lead to defects in potholes and other roads, including dents, bumps, and alligator cracks. These problems could result in acute damage to vehicle suspension systems or even increase the probability of life-threatening accidents. The World Health Organization (WHO) estimates that road accidents worldwide cause 20 to 50 million injuries and approximately 1.3 million deaths annually [13,14]. In 2014, the AAA Motor Club reported that automobile damage caused by potholes alone in the United States cost 6.4 billion dollars [15].

The monitoring of road condition could result in numerous advantages from an economic, safety, and productivity point of view [16,17]. To solve the challenges of efficient and accurate detection of road anomalies, researchers are now exploring new technical solutions. For example, Seraj et al. [18] used smartphones with GPS and inertial sensors, gyroscopes, and accelerometers to study the morphology of road surfaces. They proposed a system to detect anomalies in real-time, employing wavelet decomposition and an SVM. Celaya-Padilla et al. [19] proposed a novel technique that uses GPS, accelerometer, and gyroscope sensors mounted on a car to identify road anomalies, especially speed bumps. A genetic algorithm was used to create a logistic model based on sensor data. These studies suggest that the use of smartphone sensors is a viable and affordable way to identify road anomalies. The concept is well defined, but the lack of standard data sets and effective evaluation methodologies prevents researchers from acquiring a complete picture and viewpoint on potential solutions. Without standard datasets and assessment tools, it is impossible to identify the drawbacks, benefits, and challenges of different solutions.

In response to the lack of a benchmark dataset, several solutions have been proposed. Fox et al. [20] developed three crowd-sourcing methods for detecting potholes, considering differences in vehicles, the asynchronous operation of sensors, GPS errors, and sensor noise. To address the limited availability of significant training data, the authors used CarSim (http://www.carsim.com/), a simulator program, to generate training data. Subsequently, the model performance was validated using a real-world dataset obtained by driving on city roads near the General Motors campus in Warren, MI, USA. In another study, Fox et al. [21] introduced an approach that considered road inclination and bank angle data to identify and locate potholes in multilane scenarios. González et al. [15] introduced a new dataset to compare classification algorithms for the disruption of the road surface. Using smartphones, the data set was collected using 12 cars and trucks in the city of Chihuahua, Mexico. However, the data set is no longer publicly available. Allouch et al. [13] developed an Android application to predict road quality using GPS, triaxial accelerometers, and gyroscope sensors in real-time. After data collection, a C4.5 decision tree classifier was used to classify the data. Carlos et al. [12] published Pothole Lab, a cutting-edge open-access web platform that facilitates the creation of virtual roads with customizable types and numbers of anomalies on the route. Using this dataset, they evaluated the detection approaches for road anomalies based on accelerometer readings.

Y. Chen et al. [22] suggested an approach with two phases: road anomaly segmentation and detection. During the segmentation phase, road anomaly subsequences were recovered from the sensor data using a piecewise aggregate approximation model. Scale-invariant characteristics were learned from short local segments during the detection phase to improve accuracy. Three datasets were used in their investigation, and data collection was performed using an Android application that used video, GPS, and acceleration measures. The pothole laboratory produced the second data set used in the study [12]. They created the third dataset using the CarSim simulator, according to the methodology of the reference study [20].
On the other hand, Zheng et al. [23] presented a querying and recompiling approach for road anomaly detection that requires less data. To validate the model, they used two datasets. The initial data set used was obtained from the research on road anomaly detection by Carlos et al. [12]. To enhance the authenticity of the approach, they added anomalies to existing road data based on the proportions of other datasets found in [13,15]. Following the research presented in [20], they also used the CarSim simulator to obtain the second data set. In another study, Carlos et al. [24] introduced a more intelligent method for characterizing potholes and speed bumps. They used regression, classification, and learning-to-rank techniques. To conduct this study, they collected the data set driving across Chihuahua, Mexico, in 2018. Similarly, Zheng et al. [25] introduced a technique to identify road anomalies by comparing data windows of different lengths using the dynamic time warping (DTW) method on acceleration data. This research used a similar data set from their previous study [22].

Researchers have also conducted numerous studies on the identification of driving behavior. Zylius et al. [26] introduced an approach for classifying driving styles using data from a 3-axis accelerometer (G-sensor). The objective was to detect both safe and aggressive driving behaviors. Light vehicle data were collected for driving modes and several characteristics were examined. Carlos et al. [27] proposed a second-order representation utilizing a bag-of-words technique to model accelerometer timestamps related to aggressive driving maneuvers. Binary and multiclass classification scenarios were used to analyze this representation. They used three data sets in their research. Initially, they used the Žylius Data Set [26], consisting of triaxial accelerometer readings in normal and aggressive driving scenarios. Second, they used the Ferreira et al.[28] dataset. This data set was appropriate for multiclass classification because it included various aggressive driving events divided into seven categories. Third, the authors presented their dataset, which contains instances of single aggressive driving maneuvers.

**Table 1** summarizes all the state-of-the-art data sets, indicating that most data sets in this field are based on simulated data. Some researchers collected custom datasets from the real world. However, most of them are no longer publicly accessible. There was no international authority in charge of the data collection and sharing process. A new global database was developed to provide essential data for researchers to overcome the absence of a benchmark dataset.

*Table 1. Summary of the state-of-the-art datasets.*

| Reference | Dataset | Dataset Type | Availability | Recorded Parameters |
|---|---|---|---|---|
| Fox et al. [20] | CarSim (Training), Custom (Testing) | Simulated | Available | Distance traveled, Vehicle Speed, Accelerometer, GPS |
| Fox et al. [21] | CarSim (Training), Custom (Testing) | Simulated | Available | GPS, Accelerometer, Vehicle Speed, |
| Gonzalez et al. [15] | Custom | Real-world | Not available | Accelerometer |
| Allouch et al. [13] | Custom | Real-world | Not available | Accelerometer, Gyroscope, GPS |
| Carlos et al. [12] | Pothole Lab | Simulated | Not accessible | Accelerometer |
| Chen et al. [22] | Custom, Pothole Lab, CarSim | Real-world, Simulated | Fully not available | GPS, Accelerometer, Video |
| Zheng et al. [23] | Pothole Lab, CarSim | Simulated | Fully not available | Distance traveled, Vehicle Speed, Acceleration, GPS |
| Carlos et al. [24] | Custom | Real-world | Not accessible | Accelerometer |
| Zheng et al. [25] | Custom, Pothole Lab, CarSim | Real-world, Simulated | Fully not available | GPS, Accelerometer, Vehicle Speed, and Velocity |
| Zylius et al. [26] | Custom | Real-world | Not available | Accelerometer |
| Carlos et al. [27] | Custom | Real-world | Not available | Accelerometer, Magnetometer, Gyroscope, Linear Accelerometer |
| Proposed Dataset | Custom | Real-world | Available | Accelerometer, Gravity, Gyroscope, Orientation, Magnetometer, GPS, Total Acceleration, Uncalibrated Magnetometer, Uncalibrated Gyroscope, Uncalibrated Accelerometer |

In the past decade, the growth of autonomous vehicles has become a central research topic in the automobile industry, emerging as an auspicious technological achievement. Several states and nations passed laws to conduct tests after Nevada, USA, which allowed autonomous cars on public roads in June 2011 [29]. Recently, the advancement of artificial intelligence (AI) and Internet of Things (IoT) technologies has led to significant improvements in autonomous driving. This technology has become more intelligent in accurately interpreting real-world surroundings, quickly analyzing sensor data, and independently making complicated decisions. Despite its amazing potential, the rapid advancement of autonomous driving technologies has also been hampered by new obstacles [30]. Before the widespread implementation of autonomous vehicles in real-world traffic scenarios, it is vital to ensure the security and reliability of these vehicles through several technical measures [31]. The data set collected through several types of sensors can also play a crucial role in advancing autonomous driving technology.

This article provides the fundamentals for an in-depth study of the driving behaviors and road anomalies data set, highlighting the importance of this data set to encourage future research. Section 2 presents a detailed description of the data collection approach using smartphone sensors, demonstrating the systematic acquisition of driving behaviors and road anomalies data. Section 3 provides a detailed analysis of the collected data records. Section 4 focuses on the technical validation procedures to ensure the reliability of the dataset. Section 5 of the study outlines possible research options to explore using the data set.

## Methods
In this section, we present the data collection strategy and analysis. **Figure 1** outlines the comprehensive strategy, while subsequent subsections detail the experimental setup, parameter selection, route planning, data collection session, and considerations regarding data storage and security.

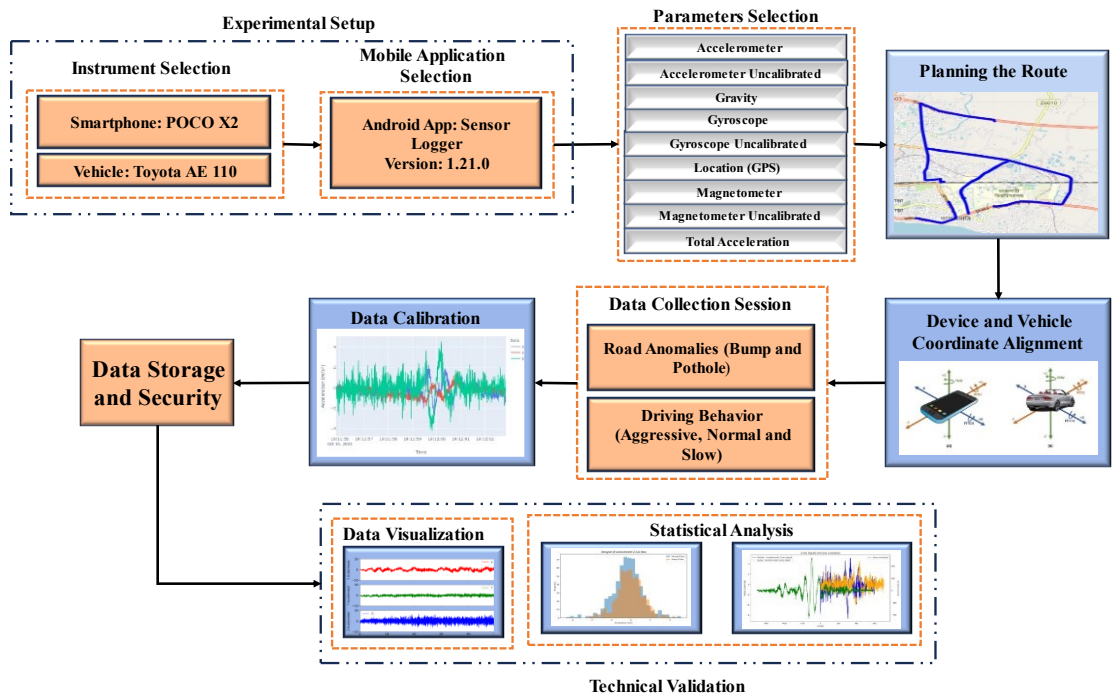

*Figure 1. Overview of the data collection methodology.*

**Experimental Setup**

For this study, we utilized a Xiaomi Poco X2 smartphone running Android 4.0, with the Sensor Logger app version 1.21.0, available on Google Play. Due to constraints in the Android API, direct access to the smartphone's sensors is not permitted. Therefore, data collection was conducted exclusively through the Sensor Logger app, which effectively manages sensor data within the limitations of the API. A single device model, the Poco X2, was chosen to ensure consistency and minimize variability in sensor performance and data quality.

A structured process was employed to evaluate the noise levels of the selected android device Poco X2, encompassing data collection and Allan variance measurement. The primary goal was to determine if the noise characteristics of the device conform to acceptable standards. At first, we recorded accelerometer data from the mobile phone over a specified duration. The phone's sensor captured acceleration signals in three orthogonal axes—x, y, and z—. Allan Variance Measurement was then conducted on the collected accelerometer data to evaluate the noise characteristics. The measurement process included essential steps:

- Calculation: Allan variance was computed across various averaging times (τ), providing insights into the noise characteristics of the device. This calculation allows for examining how variance changes with different averaging periods, categorizing the results into different types of noise: White Noise, Flicker Noise, and Random Walk.
- Analysis: The Allan variance results were analyzed to determine the noise levels for each noise type. These results were compared against predefined acceptable noise levels of IMU (Inertial Measurement Unit) device to evaluate whether the device's performance meets the established standards.

*Table 2. Allan Variance Results and Noise Level Assessment of the Data Collection Device.*

| Axis | Noise Type | Measured Noise Level | Acceptable Noise Level | Assessment |
|---|---|---|---|---|
| x | White Noise | 0.07 | White Noise: 0.05 - 0.10 | Acceptable |
|   | Flicker Noise | 0.07 | Flicker Noise: 0.05 - 0.10 | Acceptable |
|   | Random Walk | 0.02 | Random Walk: 0.01 - 0.05 | Acceptable |
| y | White Noise | 0.08 |  | Acceptable |

|   | Flicker Noise | 0.08 |   | Acceptable |
|   | Random Walk | 0.04 |   | Acceptable |
|   | White Noise | 0.05 |   | Acceptable |
| z | Flicker Noise | 0.05 |   | Acceptable |
|   | Random Walk | 0.02 |   | Acceptable |

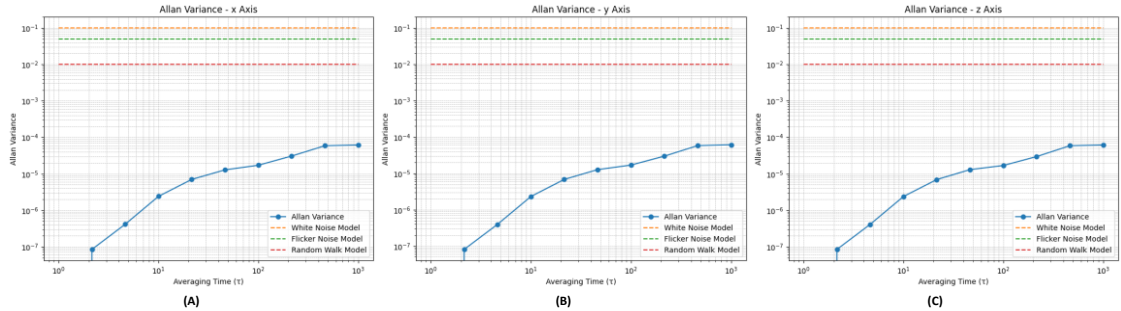

*Figure 2. Allan variance across x, y, and z-axis of the accelerometer.*

**Table 2** displays the noise levels for each axis of the mobile phone accelerometer. On the x-axis, the measured white noise level is 0.07, which falls within the acceptable range of 0.05 to 0.10, and thus is considered acceptable. Similarly, the flicker noise and random walk levels of 0.07 and 0.02, respectively, are also deemed acceptable, as they lie within their respective acceptable ranges. On the y-axis, the noise levels for white noise, flicker noise, and random walk are 0.08, 0.08, and 0.04, respectively, all within acceptable ranges and thus considered acceptable. For the z-axis, the measured values for white noise, flicker noise, and random walk are 0.05, 0.05, and 0.02, which also fall within the acceptable noise ranges, confirming their acceptability. Overall, the device exhibits acceptable noise levels across all axes and noise types, indicating that it performs well within the defined acceptable limits. **Figure 2** displays the Allan variance plot.

The Sensor Logger app, which is available for both iOS and Android, facilitates cross-platform data collection and ensures a consistent user experience across these platforms. Although there are differences in units and coordinate frames between iOS and Android, the app's "Standardize Units & Frames" feature addresses these discrepancies, enabling reliable comparisons of data collected from various devices (https://github.com/tszheichoi/awesome-sensor-logger/blob/main/CROSSPLATFORM.md) This feature is crucial for maintaining data consistency and minimizing potential discrepancies. The app has been downloaded by over 50,000 users (https://play.google.com/store/apps/details?id=com.kelvin.sensorapp&hl=en), highlighting its widespread acceptance and reliability.

To substantiate the robustness of the Sensor Logger app, a cross-device validation was conducted by collecting accelerometer data from both a Poco X2 and an iPhone 15. This validation aimed to determine if the data collected across different smartphone models using the app remained consistent. The app's "Standardize Units & Frames" feature was utilized to align measurement units and coordinate frames, facilitating device comparability. After collecting the data, accelerometer readings from both devices were processed to eliminate artifacts and normalize the signals. A visual plot was generated to compare the data patterns from the Poco X2 and the iPhone 15. The analysis of **Figure 3** plot revealed that the data from both devices exhibited similar patterns and noise levels, demonstrating that the Sensor Logger app effectively standardizes measurements. This result supports the conclusion that the dataset collected using the Poco X2 is robust and reliable, with the app ensuring consistent data across different smartphone models. The provided figure visually reinforces this consistency, illustrating the app's capability to produce reliable and comparable data.

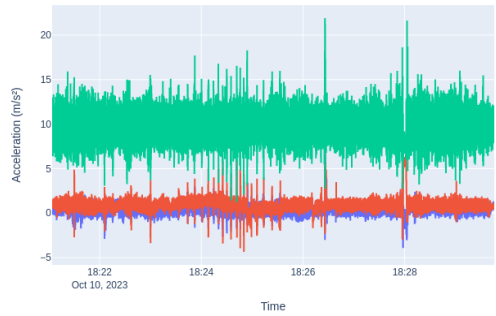
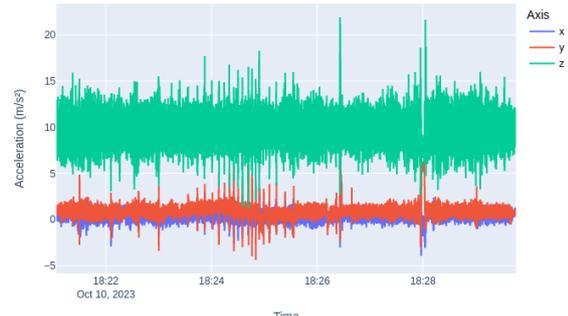

*Figure 3. Comparison of accelerometer data patterns between the Poco X2 and iPhone 15, illustrating similar trends and noise levels*

The data collection process was carried out on a 1995 Toyota SE Saloon vehicle, commonly called the Toyota AE 110 or Toyota 110. The smartphone was securely mounted inside the vehicle on the dashboard using a high-quality suction cup mount equipped with a strong adhesive base for data collection. This mounting device was selected for its reliability in maintaining a stable connection to the dashboard while minimizing vibrations that could affect sensor readings. The interior placement of the device minimized exposure to external environmental factors like sunlight and temperature variations, which can impact sensor accuracy. The smartphone was positioned centrally on the dashboard to ensure optimal alignment of the sensors with the car's primary axes (x, y, z) and to capture data from a consistent viewpoint. This central location also facilitated uniform exposure to road conditions and minimized potential interference from other vehicle components. The attachment method was rigorously tested on a variety of road surfaces before the experiments to confirm the phone's secure placement and address any potential movement risk during data collection. **Figure 4** provides a visual representation of the smartphone's exact positioning and orientation on the dashboard within the vehicle's interior, illustrating the stable and secure installation used during data collection. This setup was crucial for accurately capturing the vehicle's motion and orientation during the data collection process.

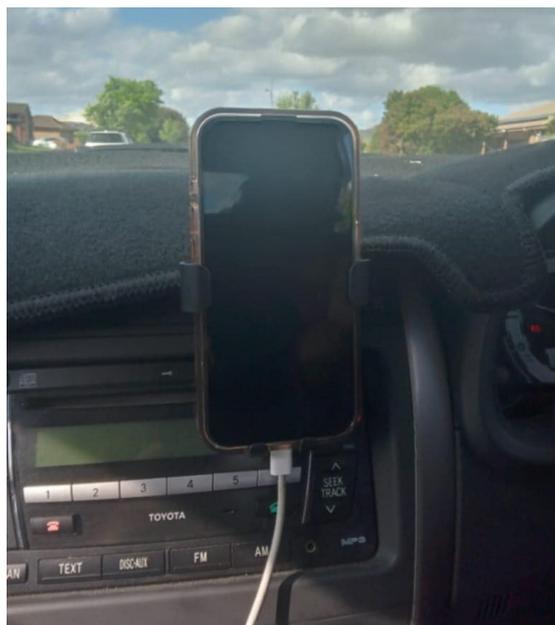

*Figure 4. Visual Representation of smartphone's installation process.*

**Parameter Selection**

Various built-in sensors, such as accelerometers, magnetometers, gyroscopes, lights, and temperature sensors, are accessible via the smartphone sensor framework [32]. These sensors fall primarily into hardware-based and software-based categories. Hardware sensors are used to determine environmental factors, motion, and orientation. When data are from hardware-based sensors, software-based sensors can dynamically compute values, including linear acceleration, orientation, and gravity. However, these sensors can be broadly classified into three groups: motion, environmental, and position sensors [33]. Motion sensors can detect movements such as rotation (roll, which is front-to-back; pitch, which is side-to-side; and yaw, which is up-down), as well as shaking and vibration of a device. Position sensors provide the location of an object relative to a local coordinate system. When the data obtained from the accelerometer and the geomagnetic field sensor are incorporated, the exact location of the device can be determined. Environmental sensors can assess a multitude of parameters, such as humidity, illumination, atmospheric pressure, and temperature.

**Route planning**

The data collection process was conducted in Rajshahi, a major city in Bangladesh. Before starting the data collection session, specific zones in the city were delineated where road anomalies such as bumps and potholes were detected. A comprehensive catalog was created containing each selected area's longitude and latitude coordinates. The data collectors then strategized and created a route plan to travel through these selected areas. **Figure 5** shows the route plan with blue dotted markers representing the data collection points on the map.

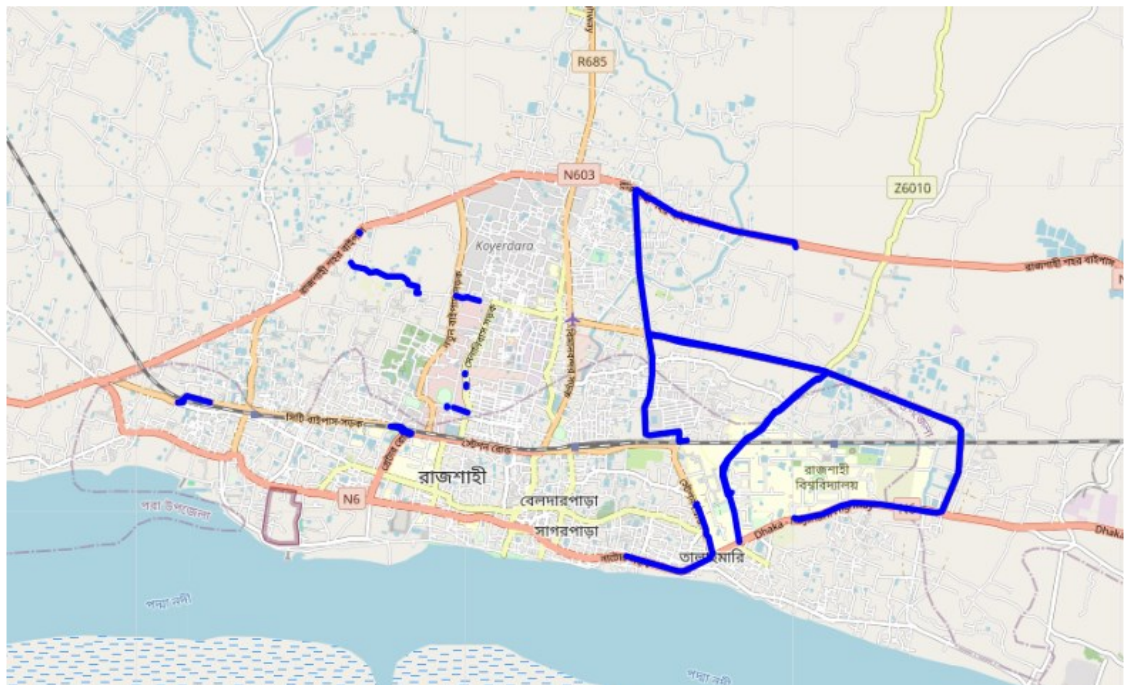

*Figure 5. Route planning (blue lines manually put to represent the data collection point).*

**Alignment of Smartphone and Vehicle Coordinate Systems**

Smartphone sensors utilize a conventional 3-axis coordinate system to represent data values, typically defined in relation to the device's screen. When the smartphone is positioned horizontally, the x-axis corresponds to the horizontal direction from left to right, the y-axis runs vertically from bottom to top, and the z-axis points perpendicular to the screen's face. As depicted in **Figure 6**, the smartphone's coordinate system was carefully aligned with the vehicle's coordinate system. This alignment ensures that the x-axis of the smartphone corresponds to the vehicle's longitudinal axis, the y-axis aligns with the lateral axis, and the z-axis aligns with the vertical axis. Consequently, the z-axis of the vehicle's coordinate system accurately represents the gravitational acceleration relative to the Earth's coordinate system. This careful alignment was crucial for accurately capturing the vehicle's motion and ensuring that the data collected by the smartphone sensors correctly reflected the vehicle's dynamics.

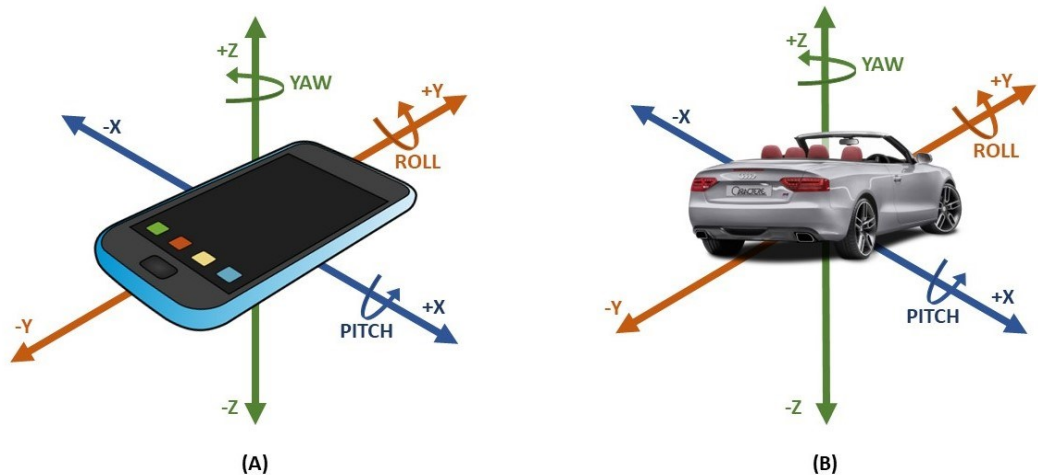

*Figure 6. Device and vehicle coordinate alignment*

Each sensor event yields multidimensional arrays of sensor readings for all motion sensors. During a single-sensor event, the accelerometer provides data on the acceleration force for the x, y, and z-axes. However, the gyroscope provides data on the rotation rate for the three coordinate axes. The gravity sensor provides a three-dimensional vector indicating the direction and strength of gravity. The orientation sensor can be used to calculate the orientation of a device. This sensor keeps track of the device's location in relation to the magnetic north pole. Changes in the magnetic field can be observed via the magnetometer sensor, which provides information about the raw-field strength for all three coordinate directions. Data values are recovered as a float array (values) together with other attributes of sensor events.

**Data Collection Session**
During the data collection process, the driver and data collectors knew the exact location of an anomaly on the road. The driver alerted the data collector to begin collecting data when the vehicle was close to an anomaly. The incident started to be recorded through the sensors on the smartphones. The data collector collected incident information after passing through the anomaly. Each event was handled using this procedure.

The details of the drivers involved in the study are as follows:
- Driver 1: Age 26, with 1 year of driving experience. This driver is relatively new to driving, with limited experience on the road.
- Driver 2: Age 32, with 4 years of driving experience. This driver has moderate driving experience, having been on the road for a few years.
- Driver 3: Age 27, with 2 years of driving experience. This driver has a couple of years of driving experience, indicating a growing familiarity with different driving conditions.

Data on aggressive driving, standard driving, and slow driving behavior were also collected using the same approach. Aggressive driving behavior is defined by a combination of actions and attitudes. Common indicators of reckless driving include exceeding the speed limit based on the present road conditions, making sudden lane changes without signaling or considering other vehicles, excessively and inappropriately using the vehicle's horn or headlights to express frustration or anger, and overlooking important traffic rules. Defining a baseline driving style for each driver has allowed deviations indicative of aggressive and slow driving to be identified. The features of baseline or standard driving behavior have included maintaining a consistent speed that aligns with road conditions and legal speed limits, and avoiding sudden and erratic accelerations or decelerations. Standard driving behavior has also involved smooth acceleration and braking, with gradual increases and decreases in speed to ensure a smooth driving experience [9]. Aggressive and slow driving styles have been recognized as deviations from standard driving behavior. Aggressive driving has been identified when a driver's behavior significantly deviates from their normal driving, including exceeding speed limits, sudden braking or intense accelerations, and making sharp turns or hazardous overtakes. These actions increase the risk of losing control and require higher safety distances. Slow driving behavior, often intentional for safety, has been characterized by consistently driving below normal and legal speed limits, ensuring smooth and gentle speed changes, maintaining a larger-than-usual distance from the vehicle ahead, and executing lane changes and turns slowly with clear signaling [10]. All these characteristics were considered when collecting data on aggressive, standard, and slow driving. The vehicle speed was intentionally reduced below the normal pace to collect data on slow driving. The vehicles had speed limits of 80-120 km/h, 40-60 km/h, and 20 km/h for aggressive driving, standard driving, and slow driving, respectively [11].

The entire data collection process involved driving a predetermined route that spans more than 30 kilometers. Safety was always the main concern during the data collection process, and all data collection activities were conducted meticulously in strict compliance with local traffic laws and regulations.

**Calibration of Sensor Data**

In the context of smartphone sensors, calibration refers to adjusting raw sensor data to account for any inherent inaccuracies or biases. The calibrated data has been adjusted to enhance accuracy and alignment with the genuine values. Uncalibrated data refers to the unprocessed output obtained directly from the sensor without any adjustments or corrections applied. Calibrated data offers more accurate and exact measurements, but uncalibrated data preserves the original sensor information, which might be useful for customized post-processing or advanced sensor fusion techniques.

Using the Sensor Logger Android app, both calibrated and uncalibrated data was collected for the accelerometer, magnetometer, and gyroscope used in the study. The application undergoes an internal calibration to eliminate problems such as sensor drift and bias. Drift refers to the gradual change in sensor readings over time, while bias indicates a consistent offset from the true value even when no input is present. When the accelerometer is not moving, it should ideally measure a value of 0 m/s² on the x and y axes and approximately 9.8 m/s² on the z-axis, which is influenced by gravity. Any deviation from these values indicates bias. Similarly, when not moving, the gyroscope should indicate no rotation in any direction. However, flaws in the manufacturing process may indicate a slight but consistent rotation, which represents the gyroscope's bias.

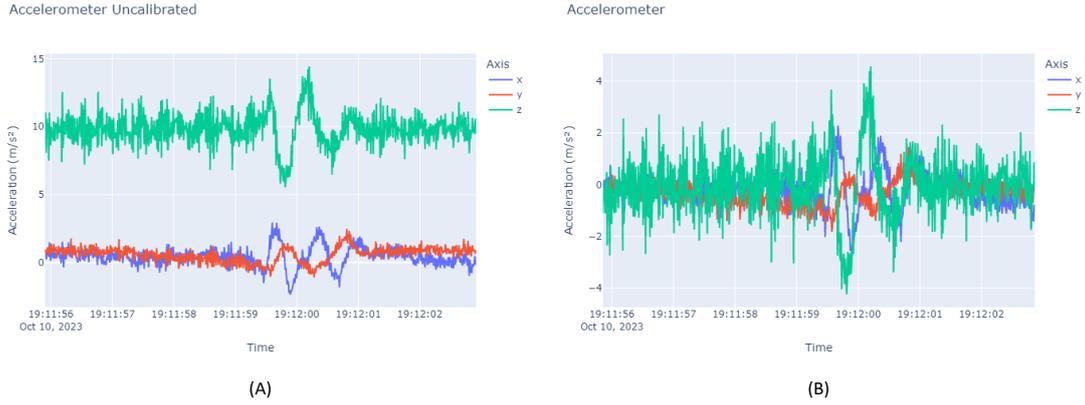

*Figure 7. Difference between calibrated and uncalibrated data. A) Accelerometer uncalibrated, B) Accelerometer Calibrated. (folder: 3. Bump)*

To calibrate the sensors, the app takes multiple measurements under controlled conditions, such as the stationary device for the accelerometer and gyroscope or placed in a non-magnetic environment for the magnetometer. These measurements are used to determine and correct for biases, also known as zero-offset errors, which are consistent deviations from the true value when the sensor should read zero. For instance, if the gyroscope is completely stationary, it should ideally indicate 0 rotation along all axes. However, due to manufacturing flaws or environmental influences, it may exhibit minor yet consistent measurements, such as -0.43 rad/s along the x-axis, indicating the presence of bias. Calibration involves calculating the offset from these measurements, which is subsequently deducted from future readings to rectify the data. For the accelerometer, calibration can be mathematically expressed as:

$$C_a = S_a - O_a \qquad (1)$$

where, $C_a$ is calibrated accelerometer output, $S_a$ is raw accelerometer measurement and $O_a$ is offset determined during calibration. After calibration, the accelerometer's x, y, and z values should ideally be reduced to zero while the device is not moving, thereby minimizing errors or deviations.

**Figure 7**, which presents data from the bump (folder: 3. Bump), illustrates the differences between calibrated and uncalibrated data from the accelerometer. In the analysis conducted, the offsets calculated from stationary uncalibrated data were found to be 0.410 m/s² for the X-axis, 0.427 m/s² for the Y-axis, and 9.814 m/s² for the Z-axis. During the same stationary period, the mean calibrated values were determined to be -0.176 m/s² for the X-axis, -0.428 m/s² for the Y-axis, and -0.059 m/s² for the Z-axis. These results, summarized in **Table 3**, demonstrate the effectiveness of the calibration process in addressing biases in sensor readings. It was confirmed that the calibration adjustments brought the readings closer to the expected values when the sensor was stationary. The visual representation in **Figure 7** effectively contextualizes how calibration minimizes biases in the sensor's output.

*Table 3. Offsets and calibrated values for accelerometer.*

| Axis | Offset (Uncalibrated) | Mean Calibrated Value |
|---|---|---|
| X | 0.410 m/s² | -0.176 m/s² |
| Y | 0.427 m/s² | -0.428 m/s² |
| Z | 9.814 m/s² | -0.059 m/s² |

In the calibration process for the magnetometer, an ellipsoid fitting approach is typically employed to compensate for hard-iron and soft-iron distortions. Initially, raw magnetometer data is collected by measuring the magnetic field at various orientations, resulting in a dataset represented as:

$$S_m = (x, y, z) \tag{2}$$

To accurately represent the magnetic field, an ellipsoid is fitted to the gathered data. The general equation for the ellipsoid is expressed as follows:

$$\frac{(x - h_x)^2}{a^2} + \frac{(y - h_y)^2}{b^2} + \frac{(z - h_z)^2}{c^2} = 1 \tag{3}$$

In this equation, $(h_x, h_y, h_z)$ denotes the center of the ellipsoid, while $a, b, c$ represent the radii along the respective axes. This fitting process allows for the identification of offsets due to sensor biases that can significantly affect measurement accuracy.

Upon successful ellipsoid fitting, the calibration of the magnetometer output can be performed using the following formula:

$$C_m = S_m - O_m \tag{4}$$

Where, $C_m$ represents the calibrated magnetometer output, $S_m$ indicates the raw magnetometer measurement, and $O_m$ is the offset determined during the ellipsoid fitting process. This method primarily focuses on correcting biases in the raw measurements, ensuring that the output more accurately reflects the true magnetic field.

In some cases, a more comprehensive calibration approach may be necessary, which incorporates both offset correction and scaling adjustments. This can be represented by the formula:

$$M = K \cdot (B - O) \tag{5}$$

Where, $M$ is the calibrated output, $B$ is the raw magnetometer measurement, $O$ is the determined offset, and $K$ is a calibration gain or scaling factor. This formula is particularly useful when the calibration requires adjustments for both bias and scaling, addressing more complex calibration scenarios.

In addition to sensor-intrinsic biases, such as zero-offset errors, external factors—including impermanence and inaccuracies in manual alignment during calibration—can introduce additional bias. During calibration, slight misalignments may occur if the device is not perfectly oriented with respect to the reference axes. For example, an unintentional tilt or rotation can lead to consistent deviations in the sensor readings. If the accelerometer is not level during calibration, it may register a small non-zero value even when stationary. Such deviations are not due to sensor-intrinsic bias but rather errors introduced during the manual alignment process. To mitigate this risk, the calibration procedure is designed to ensure precise positioning on a stable, level surface, minimizing the impact of human handling errors.

Over time, sensors may experience drift due to environmental factors such as temperature fluctuations, physical wear, or gradual shifts in the device's internal components. This drift can cause small biases to accumulate in sensor readings, even after calibration. To address this, periodic recalibration is essential for maintaining the accuracy of sensor measurements. By recalibrating the device at regular intervals or whenever deviations are detected, the bias introduced by impermanence can be effectively minimized.

The primary objective of calibration is to correct these biases. Properly executed calibration ensures that inaccuracies from manual alignment are rectified through precise positioning during the procedure. Similarly, regular recalibration addresses biases caused by sensor drift over time. By incorporating these measures, the calibration process guarantees that sensor readings accurately reflect true physical conditions, reducing errors from both intrinsic sensor factors and external influences.

Although calibration addresses bias, additional noise from random variability in sensor readings may remain. Further filtering techniques can be applied to mitigate this impact. The benefit of utilizing calibrated data lies in its immediate accuracy, essential for applications requiring precise measurements. Uncalibrated data is also valuable as it provides a comprehensive view of the sensor's raw output, which may be useful for specific applications requiring customized calibration procedures or further analysis. Our study presents both calibrated and uncalibrated data, providing flexibility in data analysis and ensuring reliable measurements and raw sensor information for comprehensive evaluation.

**Data Storage and Security**

The data collected were stored in tabular CSV format for different sensors. Ten types of sensors were used to produce the data set: accelerometer, gravity, gyroscope, orientation, magnetometer, GPS, total acceleration, uncalibrated magnetometer, uncalibrated gyroscope, and uncalibrated accelerometer. Depending on the type of application, one or more sensor outputs can be selected. An overview of all sensors used in this study is given in **Table 4,** which also shows the value range and the typical error (which are in the acceptable range).

*Table 4. List of sensors used for this study.*

| Sensor Name | | Unit | Description | Data Format | Value Range | Typical Error |
|---|---|---|---|---|---|---|
| Accelerometer | | $m/s^2$ | Measures the acceleration force. | CSV | ±19.6 m/s² | ±0.1 m/s² |
| Gravity | | $m/s^2$ | Calculates the gravitational force. [25] | CSV | ±9.8 m/s² | ±0.01 m/s² |
| Gyroscope | | $rad/s$ | Measures the rate of rotation. | CSV | ±250 rad/s | ±0.02 rad/s |
| Orientation | Euler Angles | rad | Represents the vehicle's rotation in terms of three sequential angles (roll, pitch, and yaw). | CSV | Roll, Pitch, Yaw: ±180° | ±0.5° |
| | Quaternions | --- | Represents the vehicle's rotation in a four-dimensional space, offering a more robust and gimbal-lock-free method (avoiding the loss of a degree of freedom in rotation that can occur with Euler angles). | | [-1, 1] for each component (qx, qy, qz, qw) | --- |
| Magnetometer | | $\mu T$ | Measures the strength and direction of the magnetic field. | CSV | ±1000 µT | ±0.3 µT |
| GPS | | Latitude and Longitude (°), Altitude (m) | Determines the geographical position. | CSV | Latitude/Longitude: ±180°, Altitude: ±20,000 m | ±3-5 m (position) ±15 m (altitude) |
| Total Acceleration | | $m/s^2$ | Represents the total acceleration experienced by the device. | CSV | ±19.6 m/s² | ±0.1 m/s² |
| Uncalibrated Magnetometer | | $\mu T$ | Raw data from the magnetometer sensor (without hard iron calibration) [24] | CSV | ±1000 µT | ±0.3 µT |
| Uncalibrated Gyroscope | | $rad/s$ | Measures the rate of rotation (without drift compensation). [25] | CSV | ±250 rad/s | ±0.05 rad/s |
| Uncalibrated Accelerometer | | $m/s^2$ | Raw data from the accelerometer sensor, without any modifications. | CSV | ±19.6 m/s² | ±0.1 m/s² |

Regular and systematic backups of the gathered data were also performed using Google Cloud Storage to prevent data loss and provide extra security against unanticipated events.

# Data records.
## Structure
A copy of the database has been uploaded to Figshare (The dataset will be made available upon request). The data set is divided into two diverse types of folders: Road Anomaly Data and Driving Behavior Data. There are two types of data on road anomalies. Driving behavior can be classified into three unique types. Each data folder has two primary components in the database. First, data files for road anomalies are extracted directly from the output files of smartphone sensors, such as Accelerometer.csv, Gravity.csv, and Gyroscope.csv. The overall average sampling rate for the dataset is 89.82 Hz, with observed rates ranging from 60 Hz to 99 Hz. This consistent average was maintained throughout the data acquisition process, providing a reliable dataset for both driving behavior and road anomaly analysis. These rates are aligned with typical acceptable sampling rates for smartphone sensors, which often range from 50 Hz to 100 Hz[34]. Such a sampling rate balances data resolution and storage requirements, making it suitable for capturing and analyzing dynamic events. Second, Metadata.csv provides complete information about the devices used in the study and the data collected. **Figure 8** shows the folder and file paths of the database. **Table 5** presents a summary of data statistics.

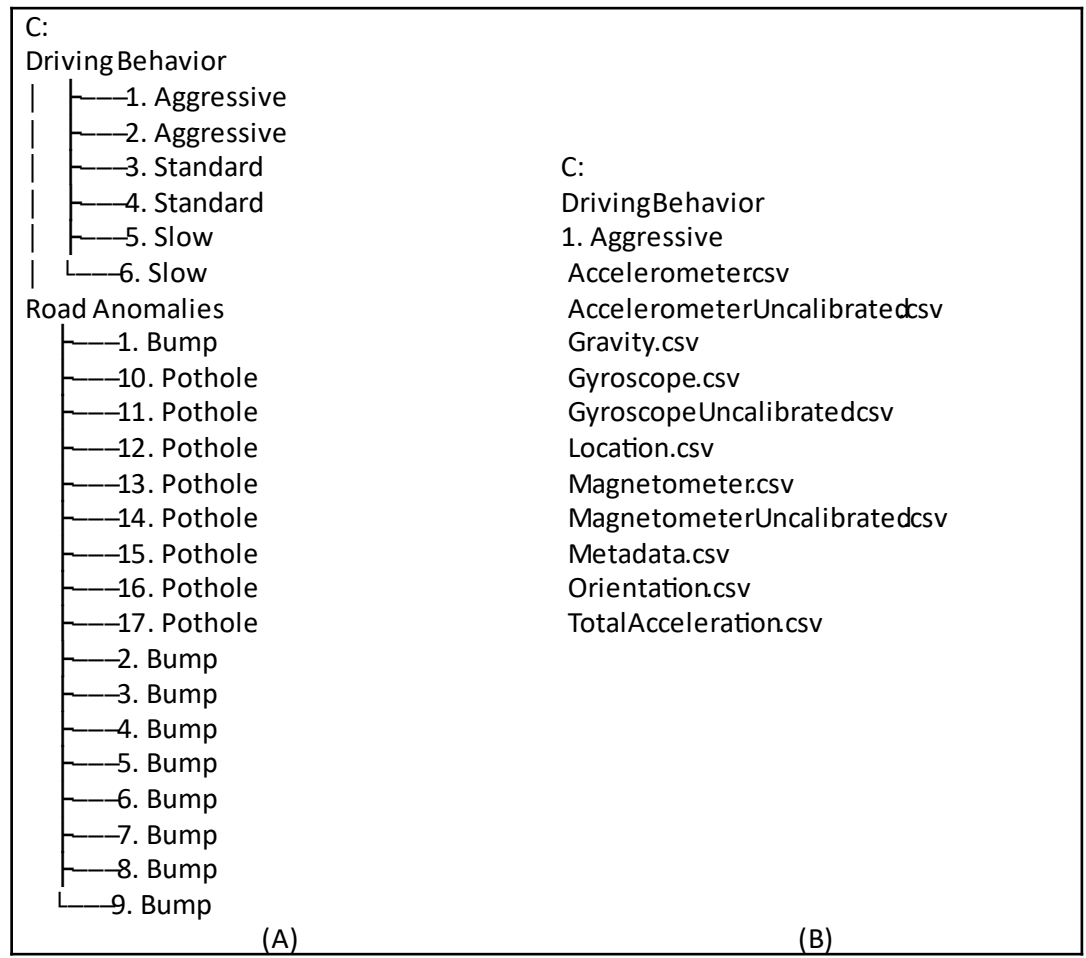

*Figure 8. (A) Path of all folders. (B) The file path of each folder.*

*Table 5. Summary of Data Statistics for Driving Behavior and Road Anomalies.*

| Category | Total Folders | Total Samples | Average Duration (s) | Standard Deviation of Duration (s) |
|---|---|---|---|---|
| Driving Behavior | 6 | 217,374 | 7,721,643.11 | 5,157,315.71 |
| Road Anomalies | 17 | 102,049 | 740,067.78 | 1,934,562.45 |

**Description of Fields.**
*Common to Accelerometer.csv, AccelerometerUncalibrated.csv, Gravity.csv, Gyroscope.csv, GyroscopeUncalibrated.csv, Magnetometer.csv, MagnetometerUncalibrated.csv, TotalAcceleration.csv.*

*Time (ns).* This term denotes the timestamp of the data point, usually in Unix time format.

*Seconds elapsed (s).* Represents the time in seconds that has passed since the initiation of data collection. It helps to measure the time interval between successive data points.

*z (m/s² or µT).* For accelerometer data, this represents acceleration along the z-axis (the vertical axis of the device), measured in meters per second squared ($m/s^2$). For magnetometer data, it indicates the magnetic field intensity along the z-axis, measured in microtesla ($\mu T$).

*y (m/s² or µT).* For accelerometer data, this represents acceleration along the y-axis (horizontal axis of the device), measured in meters per second squared ($m/s^2$). For magnetometer data, it reflects the magnetic field intensity along the y-axis, measured in microtesla ($\mu T$) (https://developer.android.com/develop/sensors-and-location/sensors/sensors_motion).

*x (m/s² or µT).* For accelerometer data, this represents acceleration along the x-axis (longitudinal axis of the device), measured in meters per second squared ($m/s^2$). For magnetometer data, it shows the magnetic field intensity along the x-axis, measured in micro tesla ($\mu T$).

*Location.csv*

*Time (ns).* This term denotes the timestamp of the data point, usually in Unix time format. This indicates the exact time at which the corresponding location data was captured.

*Seconds elapsed (s).* This variable represents the duration in seconds that has passed since the initiation of data collection. It helps to measure the time interval between successive data points.

*Bearing Accuracy (°).* Bearing accuracy refers to the accuracy of bearing information (direction).

*Speed Accuracy (m/s).* This represents the precision of the speed information in meters per second.

*Vertical Precision (m).* Refers to the vertical location or altitude measurement accuracy, expressed in meters. This metric indicates the device's ability to provide precise elevation data relative to a specific reference point.

*Horizontal Accuracy (m).* The precision of the geographical position is measured (in meters) in terms of latitude and longitude. It measures the level of accuracy in determining the exact location of the device on the Earth's surface.

*Speed (m/s).* The velocity of the device at the recorded position is denoted and measured in meters per second.

*Bearing (°).* The compass direction is measured in degrees, with 0° representing the north. Provides information on the position and direction of the device at the recorded location.

*Altitude (m).* The vertical distance above a specific reference point is usually measured in meters. This parameter denotes the elevation of the device relative to sea level.

*Longitude (°).* This value precisely indicates the geographical coordinates of the device's position on the Earth's surface with the east-west axis. It is expressed in decimal degrees.

*Latitude (°).* This parameter precisely indicates the device's geographical position on the Earth's surface along the north-south axis. It is also expressed in decimal degrees (https://developer.android.com/reference/android/location/Location).

*Orientation.csv*:

*Time (ns).* This term denotes the timestamp of the data point, usually in Unix time format. This indicates the exact time at which the corresponding location data was captured.

*Seconds elapsed (s).* This variable represents the duration in seconds that has passed since the initiation of data collection. It helps to measure the time interval between successive data points.

*Qz (unitless).* stands for the quaternion's z-component. The device's Three-dimensional orientations are represented mathematically by quaternions.

*Qy (unitless).* stands for the y-component of the quaternion. This term helps to indicate the device's orientation along with qz and qx.

*Qx (unitless).* Denotes the x-component of the quaternion.

*Qw (unitless).* stands for the quaternion's w-component. This parameter completes the definition of the orientation along with qz, qy, and qx.

*Roll (rad).* Represents the angle of rotation about the x-axis. This is the angle formed by a plane perpendicular to the ground and a plane perpendicular to the screen of the device (https://developer.android.com/develop/sensors-and-location/sensors/sensors_position).

*Pitch (rad).* The pitch angle represents the rotation of the y-axis. This is the angle formed by a plane parallel to the ground and a plane parallel to the screen of the device.

*Yaw (rad).* Represents the rotation about the z-axis. It indicates the angle between the magnetic field to the north and the current compass orientation. The yaw of the device is 0 degrees when its top edge faces magnetic north, 180 degrees when it faces south, 90 degrees when the top edge faces east, and 270 degrees when it faces west.

## Technical Validation

To ensure the accuracy and reliability of the dataset, several key issues related to data creation, sensor reliability, and further use of the data were meticulously addressed. The primary objective was to document the entire data collection and preparation process, ensuring that the dataset is robust and reliable for further analysis. The technical limitations and costs associated with integrating physical sensors into automobiles are acknowledged. Financial factors often restrict the accessibility of data. Therefore, virtual sensors were emphasized in this situation to overcome financial problems.

Particular attention was paid to the procedures followed to ensure a reliable dataset. Since smartphone sensors are the only tools used to collect data, there is a high chance that sensor malfunctions can affect data quality. The data set underwent a rigorous cleaning process to ensure its validity and efficacy for analysis. To fix possible technical problems or errors in the dataset, duplicate entries were found and eliminated. Consequently, the issue of missing values was addressed using two steps: first, any rows with missing values were eliminated to ensure the accuracy of the data, and second, the missing values in the columns were replaced with the mean values.

However, beyond cleaning, a comprehensive approach was taken to validate the data through visualization and statistical analysis. All statistical analyses were conducted using Python, leveraging libraries such as NumPy for numerical operations, SciPy for statistical tests, and Matplotlib for data visualization. These tools were instrumental in performing descriptive analysis, T-tests, and visual assessments to validate the dataset's robustness and ensure the accuracy of road anomaly and driving behavior detection. Below is a detailed breakdown of the technical validation steps undertaken:

**Road Anomalies Detection**

To ensure the accuracy and reliability of the dataset, a detailed technical validation was conducted, focusing on the analysis of sensor signals during road anomaly events, such as bumps and potholes. This validation process included plotting sensor data to visually assess the signals' integrity, followed by a rigorous statistical analysis. Key metrics such as mean, variance, and standard deviation were calculated, and a T-test was performed to confirm the statistical significance of the differences between the anomalies. Additionally, cross-correlation analysis was employed to examine the temporal relationship between bump and pothole signals. This thorough validation confirms the dataset's robustness, making it a valuable resource for further research and practical applications in road safety and vehicle dynamics.

*Data Plot Visualization*

All data signals were plotted methodically to evaluate the accuracy of the sensors throughout the quality control stage. It was essential to monitor the accuracy of the sensors regularly. Irregularities on the road surface, such as bumps and potholes, can result in abrupt variations in vehicle acceleration. When a vehicle hits a bump, the accelerometer might detect a brief spike in acceleration. However, potholes can cause rapid reduction in acceleration when the vehicle enters the hole [19]. Road anomalies can also result in variations in vehicle orientation, manifesting as spikes or fluctuations in gyroscope data, and sudden changes in altitude temporarily affect GPS data [35]. **Figures 9** and **Figure 10** shows all the sensor signals for a bump (folder: 3. Bump) and a pothole (folder: 13. Pothole) respectively.

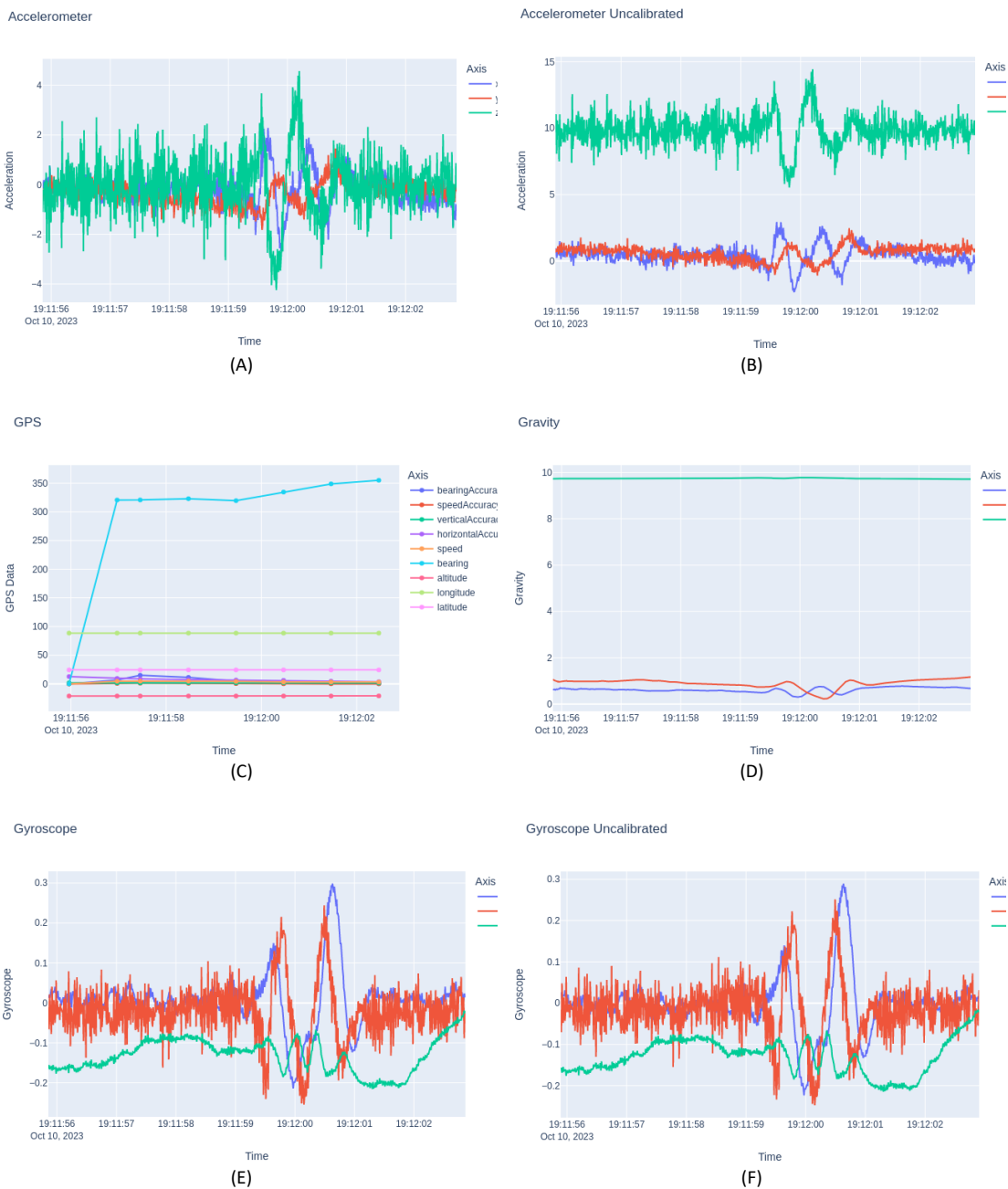

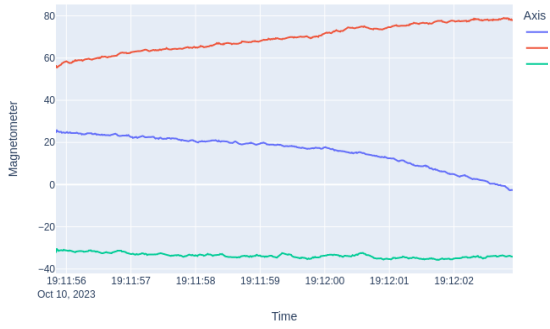

(G)

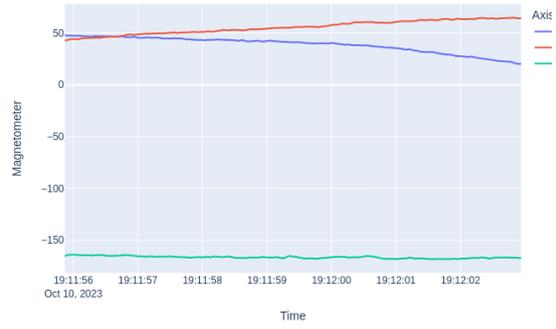

(H)

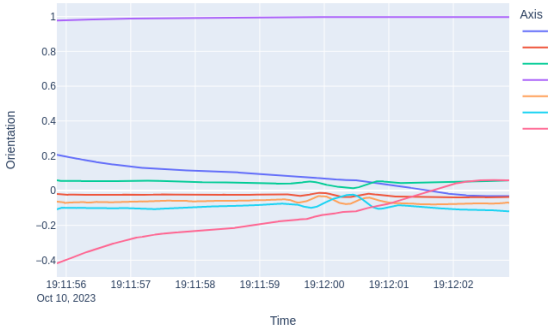

(I)

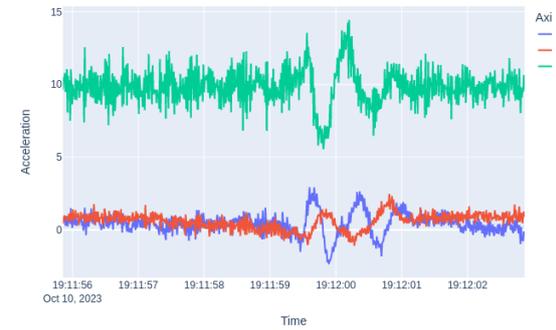

(J)

*Figure 9. Data plot of a bump for all sensors. (A) Accelerometer Calibrated, (B) Accelerometer Uncalibrated, (C) GPS, (D) Gravity, (E) Gyroscope Calibrated, (F) Gyroscope Uncalibrated, (G) Magnetometer Calibrated, (H) Magnetometer Uncalibrated, (I) Orientation, (J) Total Acceleration.*

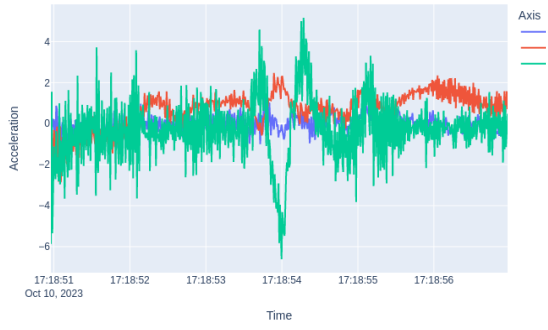

(A)

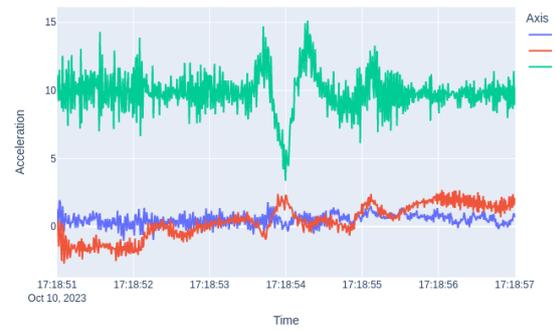

(B)

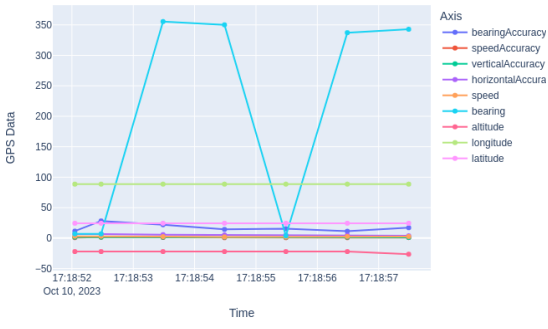

(C)

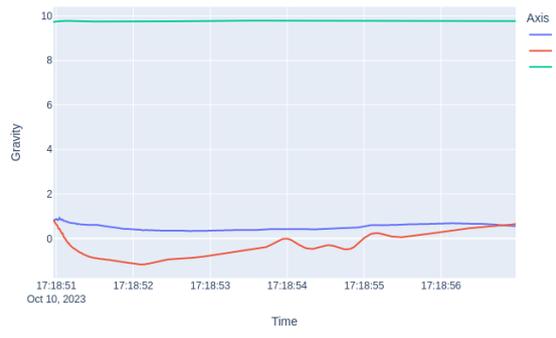

(D)

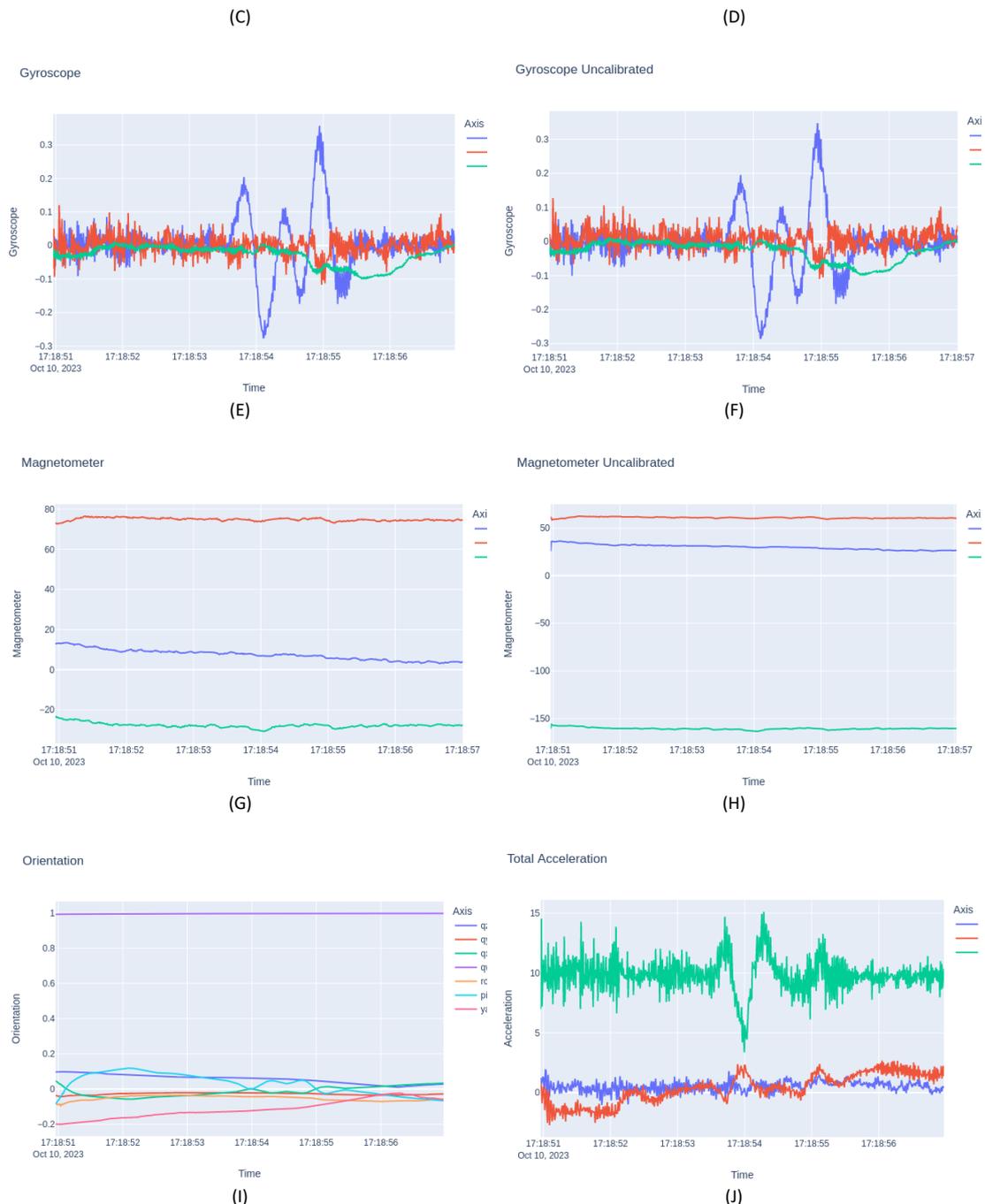

*Figure 10. Data plot of a pothole for all sensors. (A) Accelerometer Calibrated, (B) Accelerometer Uncalibrated, (C) GPS, (D) Gravity, (E) Gyroscope Calibrated, (F) Gyroscope Uncalibrated, (G) Magnetometer Calibrated, (H) Magnetometer Uncalibrated, (I) Orientation, (J) Total Acceleration.*

### *Statistical Analysis*

To substantiate the distinction between bump and pothole road anomalies, a comprehensive statistical analysis of z-axis accelerometer data was conducted, encompassing descriptive statistics, T-tests, histograms, boxplots, and cross-correlation assessments. Descriptive statistics for both bump and pothole accelerometer data are summarized in **Table 6**. The pothole data shows a higher mean, variance, range, and standard deviation compared to the bump data, indicating more substantial variations in acceleration during pothole events. A two-sample T-test was performed to assess the statistical significance of these differences, yielding a T-statistic of -3.3488 and a p-value of 0.0008. The low p-value confirms that the differences between bump and pothole data are statistically significant, leading to the rejection of the null hypothesis. Histograms, displayed in **Figure 11(B)**, reveal overlapping distributions but a broader range of acceleration values for potholes compared to bumps. Boxplots, shown in **Figure 11(A)**, further illustrate greater variability and a larger interquartile range for pothole data. This comprehensive validation through statistical analysis and visualizations confirms the distinct nature of bump and pothole accelerometer signals, reinforcing the reliability of accelerometer-based methods for detecting and distinguishing road surface defects.

*Table 6. Statistical Analysis of Accelerometer Z-Axis Data of Pothole and Bump.*

| Metric | Pothole | Bump | T-Test Result |
|---|---|---|---|
| Mean | - 0.3168 | - 0.0345 | --- |
| Variance | 2.4729 | 1.7980 | --- |
| Range | 11.7764 | 8.8127 | --- |
| Standard Deviation | 1.5725 | 1.3409 | --- |
| T-Statistic | --- | --- | -3.3488 |
| P-Value | --- | --- | 0.0008 |

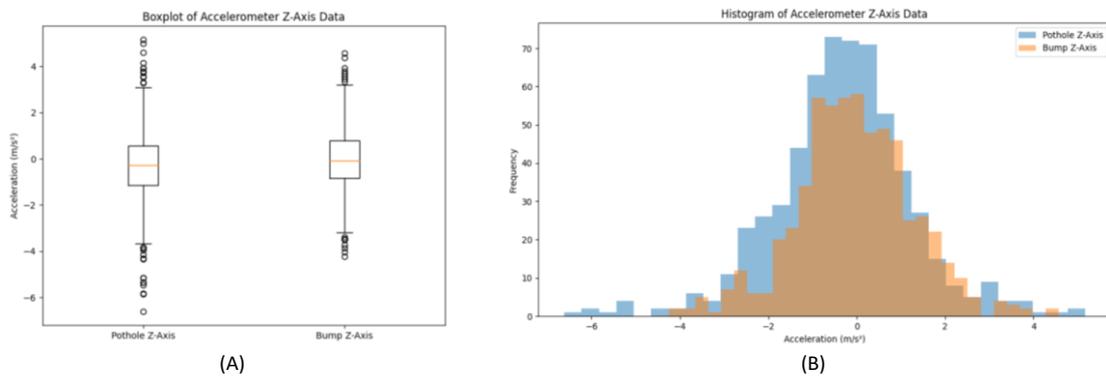

(A)        (B)

*Figure 11. Statistical analysis of bump versus pothole using z-axis accelerometer data: (A) Boxplot analysis illustrating the distribution and variance in acceleration data for bumps and potholes, and (B) Histogram analysis showing the frequency distribution of z-axis accelerations for each event type.*

## *Multivariate Sensor Analysis for Road Anomaly Detection*

For the inclusion of multiple sensors, a multivariate analysis was conducted using data from accelerometer, gyroscope, magnetometer, and orientation sensors to distinguish between road anomalies, specifically bumps and potholes. This approach leverages data from all available sensors to validate their combined effectiveness in anomaly detection, going beyond the accelerometer-only analysis initially presented.

Each sensor dataset, including both calibrated and uncalibrated data (where applicable), was merged based on a common timestamp (seconds_elapsed). To reduce dimensionality while retaining key variance, Principal Component Analysis (PCA) was applied, obtaining three principal components (PCA1, PCA2, and PCA3) that encapsulate the majority of the variance across the combined sensor features. The extracted principal components were then used as features for statistical analysis to evaluate if significant differences exist between bump and pothole events. For each principal component, two statistical tests were performed:

- **Mann-Whitney U Test** – to detect significant differences in the median values of each component between bump and pothole events.
- **Kolmogorov-Smirnov (KS) Test** – to assess differences in the distribution shapes of the components between the two event types.

The results showed statistically significant differences for each principal component, indicating that PCA1, PCA2, and PCA3 can effectively differentiate between bumps and potholes when using multivariate sensor data. **Table 7** presents the test results for each component.

*Table 7. Mann-Whitney U Test result and Kolmogorov-Smirnov Test result for each PCA.*

| Principal Component | Mann-Whitney U Test | p-value | Kolmogorov-Smirnov Test | p-value |
|---|---|---|---|---|
| PCA1 | U_stat = 386,952.0 | 1.12e-158 | KS_stat = 0.935 | 7.39e-305 |
| PCA2 | U_stat = 392,129.0 | 8.08e-168 | KS_stat = 0.947 | 1.91e-316 |
| PCA3 | U_stat = 230,861.0 | 3.73e-04 | Not significant | Not significant |

The PCA plot in **Figure 12** illustrates the distribution of bump and pothole events across the three principal components (PCA1, PCA2, and PCA3). In this scatterplot, distinct clusters of points for bump and pothole events reflect the separability provided by the combined sensor data. Additionally, 3D density plots for each significant feature from the PCA analysis were created to visualize the distributional differences, providing a more detailed view of density variations across events.

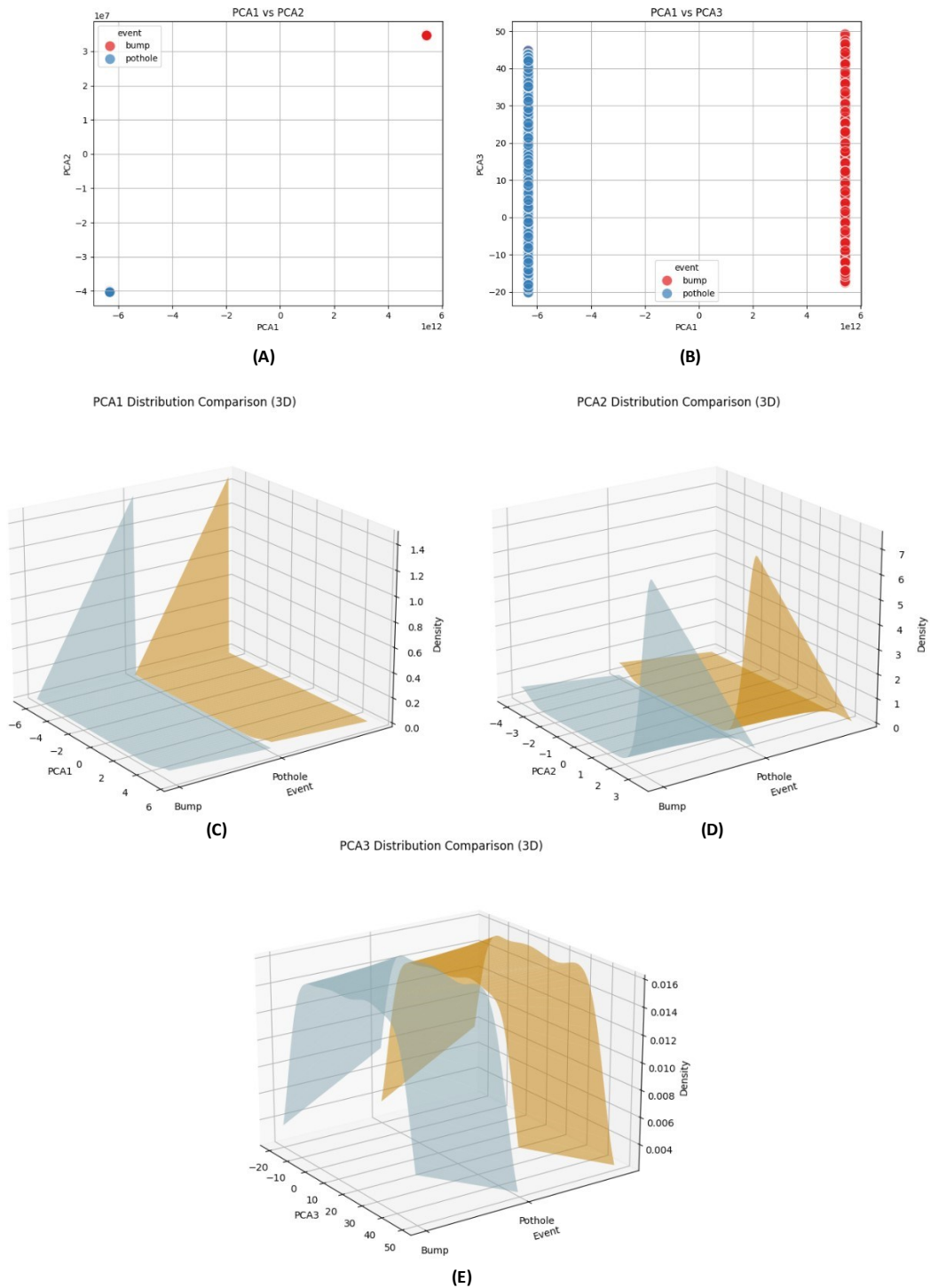

*Figure 12. PCA plot illustrating the differences between bump and pothole events across three principal components: (A) Scatter plot of PCA1 vs PCA2, (B) Scatter plot of PCA1 vs PCA3, (C) PCA1 Distribution Comparison, (D) PCA2 Distribution Comparison, (E) PCA3 Distribution Comparison.*

These findings validate that data from multiple sensors can effectively distinguish road anomaly types, supporting the inclusion of multivariate sensor data in anomaly detection applications.

**Analyzing Driving Behavior**

To evaluate driving behaviors using smartphone sensors, an extensive approach that integrates data from various sensors is employed to distinguish between aggressive, standard, and slow driving patterns. This methodology leverages multiple sensor modalities to capture a comprehensive view of driving dynamics, enabling the identification of subtle differences in driving behavior.

In order to ascertain the statistical significance of the observed differences among the driving patterns, a series of detailed analyses were conducted. These analyses included calculating the mean, variance, range, and standard deviation for each driving behavior, providing a robust statistical description of the data. Additionally, inferential statistical tests, such as the t-test, were applied to determine whether the differences between driving behaviors were statistically significant. This methodology leverages multiple sensor modalities to capture a comprehensive view of driving dynamics, enabling the identification of subtle differences in driving behavior.

*Data Plot Visualization*

Accelerometer data demonstrate rapid acceleration and deceleration in aggressive driving situations, while continuously low acceleration values indicate slow driving. Gyroscope data indicate sudden fluctuations in rotational motion for aggressive driving and minimal rotation or moderate variations in orientation for slow driving. GPS data contributes to information on speed variations and adherence to speed limits. **Figure 13** shows a comparison among aggressive driving (folder: 1. Aggressive), standard driving (folder: 3. Standard), and slow driving (folder: 5. Slow) behavior based on the accelerometer signal.

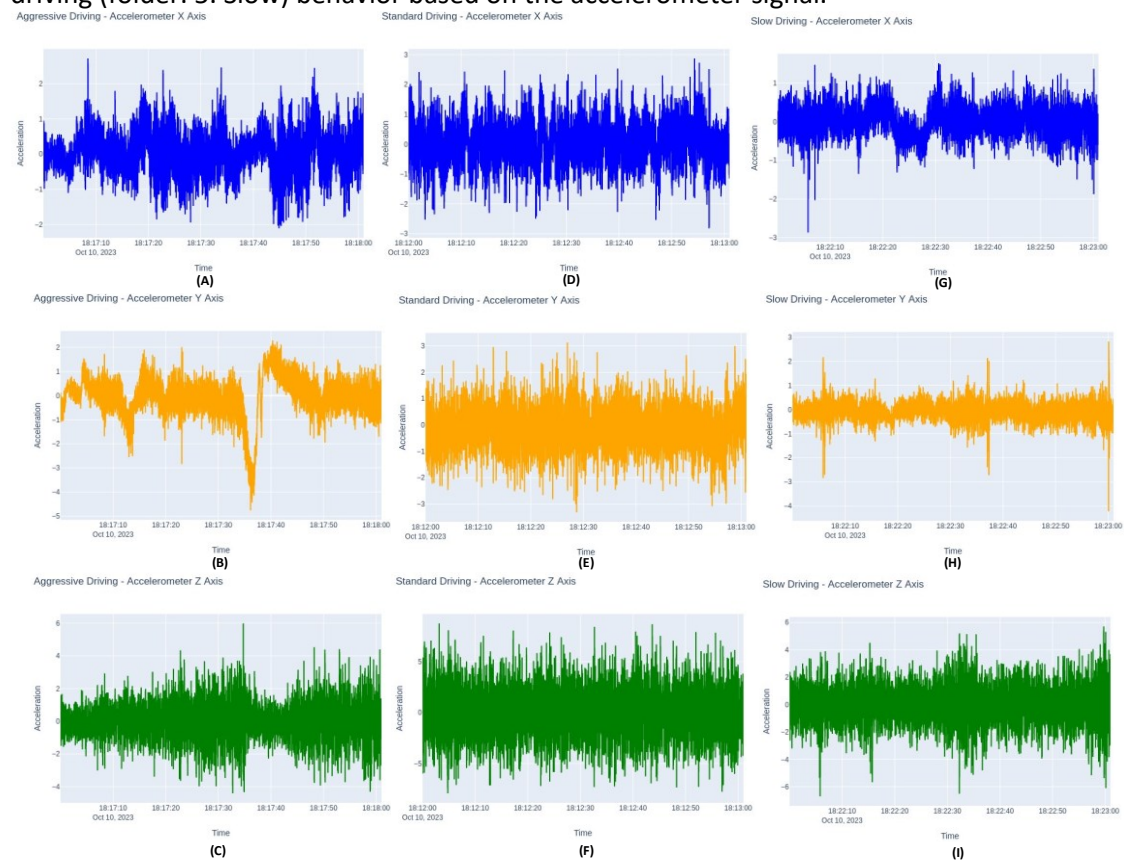

*Figure 13. Comparison among various driving behaviors based on accelerometer signals. (A) Aggressive driving – X axis, (B) Aggressive driving – Y axis, (C) Aggressive driving – Z axis, (D) Standard driving – X axis, (E) Standard driving – Y axis, (F) Standard – Z axis, (G) Slow driving – X axis, (H) Slow driving – Y axis, (I) Slow driving – Z axis.*

## Usage Notes

The dataset obtained from smartphone sensors on roadways offers a robust foundation for various research endeavors, with significant implications for real-world applications. The following highlights how this dataset can be utilized to address practical issues and enhance our understanding of road safety, infrastructure maintenance, and autonomous vehicle technology. **Table 7** illustrates the potential analyzes that can be conducted with the proposed dataset, highlighting its advantages over existing data sets.

*Table 8. Potential Analyses Enabled by the Proposed Dataset and Existing Datasets*

| Dataset | Research Opportunities with the Dataset |
|---|---|
| CarSim Dataset [20–23,25] | Pothole detection |
| Dataset by Gonzalez et al. [15] | Road anomaly detection |
| Dataset by Allouch et al. [13] | Monitoring the road condition |
| Pothole Lab [12,22,23,25] | Road anomaly detection |
| Dataset by Chen et al. [22] | Road anomaly detection |
| Dataset by Carlos et al. [24] | Characterizing potholes and speed bumps |
| Dataset by Zheng et al. [25] | Road anomaly detection |
| Dataset by Zylius et al. [26] | Identifying the driving pattern |
| Dataset by Carlos et al. [27] | Analyzing the driving pattern |
| Proposed Dataset | Road anomaly detection, Detection, and classification of driving behavior, Predictive modeling of road conditions, Safety monitoring, Urban planning and infrastructure improvements, analyzing orientation data to enhance the understanding of driver-vehicle interactions across varying conditions, and improving autonomous vehicle technology. |

**Analyzing road anomalies with accelerometer data**

A key approach to evaluating how road anomalies affect vehicle dynamics is to determine the amount of bounce a vehicle experience by measuring its vertical acceleration. Accelerometers are crucial for evaluating how road anomalies affect vehicle dynamics by measuring acceleration in various axes. When a vehicle encounters road anomalies such as bumps or potholes, the accelerometer captures both vertical acceleration (indicating bounce) and rotational movements due to the disturbances[36]. On a flat, smooth road, the accelerometer primarily measures the force of gravity, which is approximated as an acceleration of $9.81\ m/s^2$. This gravitational component can be represented as:

$$g = 9.81\ m/s^2 \qquad (6)$$

In addition to gravitational force, the accelerometer measures dynamic forces resulting from road anomalies. The total acceleration $\vec{a}_{total}$ recorded by the accelerometer is a combination of gravitational acceleration $\vec{g}$ and dynamic acceleration $\vec{a}_{dynamic}$.

$$\vec{a}_{total} = \vec{a}_{dynamic} + \vec{g} \qquad (7)$$

To accurately determine the vehicle's response to road anomalies, it is essential to isolate the dynamic acceleration $\vec{a}_{dynamic}$ by accounting for and subtracting the gravitational component. This can be achieved using:

$$\vec{a}_{dynamic} = \vec{a}_{total} - \vec{g} \qquad (8)$$

This approach allows for a comprehensive assessment of how road anomalies impact vehicle dynamics, including both vertical and rotational effects. When a vehicle hits a pothole, as shown in **Figure 14(A)**, the accelerometer records the lateral Ya and vertical Xa displacements of the vehicle, which are equivalent to the depth of the pothole. Additionally, the vehicle orientation changes, rolling to one side at an angle ɸ. On the other hand, when the vehicle hits a bump, as shown in **Figure 14(B)**, the accelerometer measures only the vertical displacement Xa, which is equal to the height of the bump. In addition, the vehicle undergoes a change in orientation, specifically when it is elevated at an angle θ [18].

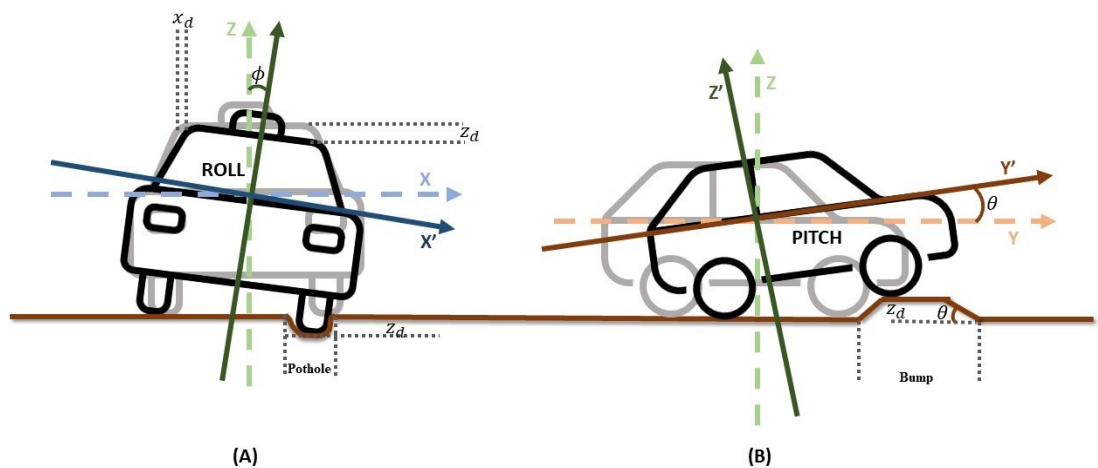

*Figure 14. Orientation of the car over road anomalies. (A) Pothole; (B) Bump*

**Behavioral analysis and safety measures**

Leveraging smartphone sensor data to analyze driving behavior offers a multifaceted approach to enhancing road safety. By integrating data from accelerometers, gyroscopes, and GPS, researchers can obtain comprehensive insights into individual driving habits, facilitating the promotion of safer driving practices. For instance, pilot programs such as the European Commission's eCall initiative employ similar sensor technologies to provide real-time feedback to drivers, effectively reducing accident rates. Additionally, applications like DriveWell (https://www.cmtelematics.com/Safe-Driving-Technology/How-It-Works/) by Cambridge Mobile Telematics utilize smartphone data to offer personalized driving feedback, which has been demonstrated to decrease risky driving behaviors. This continuous monitoring and feedback mechanism can identify hazardous driving patterns and enable proactive interventions to prevent accidents. Notable implementations include advanced driver assistance systems (ADAS), which use comparable sensor technologies to enhance driver safety and are currently integrated into various vehicles [28].

**Maintenance of road infrastructure**

Smartphone sensors such as accelerometers, gyroscopes, GPS, and magnetometers can be used to monitor road infrastructure [33]. Road anomalies can be detected and classified by analysing sensor data, indicating whether infrastructure improvements are needed. For instance, road condition monitoring systems have been developed that utilize GPS and sensor data to upload information about road conditions to central servers, which is then accessible to all application users. This technology has been successfully implemented in several smart city initiatives, such as the Road Quality Monitoring System in Singapore, which utilizes similar approaches to enhance urban infrastructure planning and maintenance.

**Enhancing the technology of autonomous vehicles**

Accurate knowledge of road anomalies is crucial for autonomous vehicles to adjust their driving characteristics for a secure journey. Autonomous vehicles use GPS and accelerometer data to assess driving behavior and traffic patterns, allowing them to optimize routes and improve traffic efficiency. Real-world applications of this technology include adaptive suspension systems in autonomous vehicles that adjust based on detected road conditions, as seen in recent advancements by companies like Waymo and Tesla. Additionally, combining smartphone sensor data with mapping information has significantly improved navigation and route optimization for autonomous vehicles. This data-driven approach aids in detecting road irregularities and enhances vehicle adaptability to dynamic road conditions, thereby improving overall safety, efficiency and performance [37,38].

## Code availability
Python code used for the curation, analysis, and validation of this dataset will be made available upon request to support reproducibility and facilitate further research.

## Acknowledgments
The authors extend their appreciation to AI tools like ChatGPT, Quill Bot, and Grammarly for their invaluable assistance in refining and improving the clarity of the writing.

## Authors' contributions
A.S.: collected data and initial draft. M.N. wrote the initial draft of the manuscript. MN: assisted with writing and drafting the manuscript and conducted data analyses. AK and JH: supervised the study and manuscript preparation and provided critical revisions. All authors have read and approved the final manuscript.

## Competing Interests
None.